\documentclass[11pt]{article}

\usepackage[final]{acl}

\usepackage{times}
\usepackage{latexsym}

\usepackage[T1]{fontenc}

\usepackage[utf8]{inputenc}

\usepackage{microtype}

\usepackage{inconsolata}

\usepackage{graphicx}
\usepackage{caption}
\usepackage{tabularx}   
\usepackage{makecell}   
\usepackage{booktabs}   
\usepackage{booktabs,tabularx,siunitx}
\usepackage[x11names]{xcolor} 
\usepackage{booktabs}
\usepackage{multirow}
\usepackage{graphicx}
\usepackage{colortbl}
\usepackage{xcolor}
\usepackage{microtype}
\usepackage{booktabs}
\usepackage{kotex}
\usepackage{bbm}
\usepackage{algorithm}
\usepackage{url}
\usepackage{soul}
\usepackage{verbatim}
\usepackage{enumitem}
\usepackage{algorithm}
\usepackage{algorithmic}
\usepackage{amsmath, amssymb}       
\usepackage{mathtools}              
\usepackage{bm} 
\usepackage{subcaption}
\usepackage{multirow}
\usepackage{lipsum}
\usepackage{booktabs}
\usepackage{multirow}
\usepackage{booktabs}
\usepackage{pifont}

\newcommand{\km}[1]{\textcolor{black}{#1}}
\newcommand{\hj}[1]{\textcolor{black}{#1}}
\newcommand{\hh}[1]{\textcolor{black}{#1}}

\title{Team MKC at CLPsych 2026: Capturing and Characterizing Mental Health Changes through Social Media Timeline Dynamics}

\author{
  Kyomin Hwang\thanks{\hspace{1pt} Equal contribution.} \quad
  Hyeonjin Kim\footnotemark[1] \quad
  Hyunho Lee\footnotemark[1] \quad
  Nojun Kwak\thanks{\hspace{1pt} Corresponding author.} \\
  Seoul National University, Seoul, Republic of Korea \\
  \texttt{\{kyomin98, peaceful1, hhlee822, nojunk\}@snu.ac.kr} \\
}

\begin{document}
\maketitle
\begin{abstract}
   \km{Recent advances in Large Language Models (LLMs) have motivated their adoption across a wide range of domains, including Artificial Intelligence (AI) for mental health. Given the growing prevalence of mental health disorders worldwide and the limited accessibility of professional care, there is an increasing demand for scalable computational approaches that can assist in early detection and continuous monitoring of psychological well-being. In this area, ongoing efforts have focused on curating domain-specific datasets and leveraging them to develop LLMs capable of supporting holistic mental health analysis. In line with this direction, we propose an LLM-based pipeline for comprehensive mental health analysis over sequentially ordered user posts, as part of the CLPsych shared task. Our pipeline offers a unified framework that jointly enables post-level assessment and user-level temporal modeling.}
\end{abstract}

\section{Introduction}

\km{Recent advances in data collection and computational power have enabled the development of Large Language Models (LLMs) capable of performing a wide range of tasks as generalist AI systems~\cite{brown2020language}. \hj{General-purpose} LLMs such as GPT~\cite{achiam2023gpt} and Claude~\cite{anthropic2025claude37} demonstrate remarkable versatility across diverse domains, from natural language understanding to complex reasoning and code synthesis. Trained on vast corpora spanning web text, scientific literature, and source code, these models acquire broad world knowledge and generalize to novel tasks with minimal task-specific supervision~\cite{kaplan2020scaling, Kim_2024_CVPR, Radford2019LanguageMA}. As the capabilities of LLMs continue to expand, there is growing interest in applying these models to specialized fields such as medical, law, and scientific discovery~\cite{xie2023darwin, hwang2026retrieval, colombo2024saullm}.}

\km{As part of these advances, there \hj{have been efforts to apply} LLMs to clinical psychology as well. \km{However, }\hj{early attempts have} faced a critical bottleneck: the scarcity of training data, which stems largely from privacy concerns surrounding sensitive personal information. To address this limitation, \citet{tsakalidis-etal-2022-identifying} \hj{released a dataset while simultaneously introducing a novel task of identifying moments of change---temporal points at which shifts in a user's mood become observable.} A growing body of subsequent work are leveraging this resource to formulate a range of tasks aimed at broader application of LLMs on mental health, with methodologies to tackle them~\cite{ali2026overview, tseriotou-etal-2025-overview, atzil_slonim_2026, tsakalidis-etal-2022-overview}.}

\km{Building on these advancements, }\km{we present an LLM-based pipeline for mental health analysis, developed as part of the CLPsych 2026 shared task~\cite{ali2026overview}, which leverages a subset of the dataset introduced by \citet{tsakalidis-etal-2022-identifying}. Specifically, our pipeline addresses five interrelated tasks: 1) assessing a user's mental state at the post level; 2) estimating the presence rate of each mental state across a user's posting history; 3) identifying Moments of Change within a chronologically ordered sequence of a user's posts; 4) summarizing the change events observed in such sequences; and 5) detecting recurrent patterns \hj{and extracting the signature of improvement and deterioration over time}. Our pipeline tackles these tasks in integration, spanning post-level assessment, user-level aggregation, and temporal modeling of mental state dynamics.}
\section{Method}

\km{In this section, we present a framework for detecting how the mental state and well-being of individuals evolve over temporally ordered sequences of social media posts. The proposed framework consists of five components: 1) post-level identification of ABCD elements and self-state, 2) self-state presence rating, 3) identification of moments of change (MOC), 4) summarization of sequences surrounding change events, and 5) identification of recurrent dynamic signatures of change across timelines. In the following subsections, we describe in detail the method employed to address each component.}

\subsection{Step 1}

\km{The first step \hj{is to classify} ABCD sub-elements~\cite{atzil_slonim_2025_mind} and self-states at the post level. Self-states are categorized into two types, adaptive and maladaptive, each comprising six subdimensions. \hj{Each post is classified into} one value from the subdimension-specific label set within each subdimension, with \textsc{None} always included as a valid option. This yields 12 independent classification targets per post, where both the number and the semantics of candidate labels vary across subdimensions, while the \textsc{None} option remains consistently available across all targets.}

\km{To address the multi-target classification problem, we employ Qwen3-4B-Embedding (Qwen Embed)~\cite{zhang2025qwen3} as a feature extractor and fine-tune it via LoRA~\cite{hu2022lora}. The \hj{model is given a} carefully designed prompt as in Figure~\ref{fig:prompt_1}. The resulting embeddings are passed to 12 independent classifiers, each trained with Cross-Entropy loss to handle its respective subdimension.}

\km{A central challenge is the severe class imbalance in the train dataset, which causes naive training to bias the model toward majority classes. We address this in two complementary ways. First, we apply inverse-frequency weighting to the Cross-Entropy loss, assigning each class a weight inversely proportional to its frequency. To prevent the extreme imbalance from inflating weights of rare classes and destabilizing training, we clip all weights to a maximum of 10. Second, since using only a subset of training data under such imbalance risks learning inadequate representations for minority classes, we adopt a 5-fold cross-validation strategy, training a separate model on each fold and ensembling the five models for final prediction.}

\subsection{Step 2}

\km{The second step builds on the outputs of Step 1 and focuses on predicting a numerical score for each self-state. Each of the two self-states, adaptive and maladaptive, is associated with a 5-point Likert scale score. The goal of Step 2 is to predict these two scores individually.}

\km{\hj{For this step}, we introduce a dedicated predictor for each self-state, which regresses a single scalar value from the embeddings produced by the feature extractor in Step 1. Each predictor is trained using a weighted Huber loss~\cite{Huber1964RobustEO}, where per-sample weights are derived from the inverse frequency of each score bin (scores 1–5) in the training set, ensuring that underrepresented score levels receive proportionally greater emphasis during optimization. The model output is produced by passing the final linear layer's logit through a sigmoid activation and linearly rescaling it to the target range [1,5]. We adopt the same 5-fold cross-validation strategy as in Step 1, training a separate model per fold and ensembling them for the final prediction. The utilized prompt is in Figure~\ref{fig:prompt_2}.}

\subsection{Step 3}

\km{In the third step, we address the task of detecting two distinct types of transitions from a temporally ordered sequence of posts: 1) \textbf{Switch}, a substantial and sudden change in well-being between two consecutive posts, and 2) \textbf{Escalation}, a gradual intensification of mood over a sequence of consecutive posts. \hj{We formulated this step as binary classification problem for both types}.} 

\km{We fine-tune the Qwen Embed using LoRA, attaching two independent linear classifier for Switch and Escalation, each trained separately to specialize in its respective transition pattern. The model operates on a sliding window of size 2, taking as input the posts at time steps $t-1$ and $t$ to predict whether a Switch or Escalation occurs at time $t$. This design reflects the local temporal dependency inherent in each task, particularly for Switch, which is defined over consecutive post pairs.}

\km{To account for class imbalance, we employ Binary Cross-Entropy with adaptive class weighting, where each label's loss weight is set inversely proportional to its frequency in the training set. Given that training on a single fold under such imbalance risks learning inadequate representations for minority classes, we adopt a 5-fold cross-validation strategy, training a separate model per fold and ensembling the five predictions for the final output. The prompt used in this step is shown in Figure~\ref{fig:prompt_3}.}

\subsection{Step 4}

\km{In the fourth step, given a sequence of posts with their associated information (\textit{e.g.}, sub-elements and self state), the objective is to generate a summary of the entire sequence. To this end, we perform Supervised Fine-Tuning on Qwen3-4B-Instruct-2507~\cite{yang2025qwen3} and Qwen3.5-4B~\cite{qwen3_5} with LoRA. During training, we provide the ground-truth annotations of each post as input, whereas at inference time we instead rely on the predictions produced by the models trained in Steps 1, 2, and 3, thereby faithfully reflecting the actual pipeline setting. The prompts used in this step are provided in Figure~\ref{fig:prompt_4_1} and Figure~\ref{fig:prompt_4_2}.}

\subsection{Step 5}

\begin{figure*}[t]
    \centering
    \includegraphics[width=1.0\linewidth]{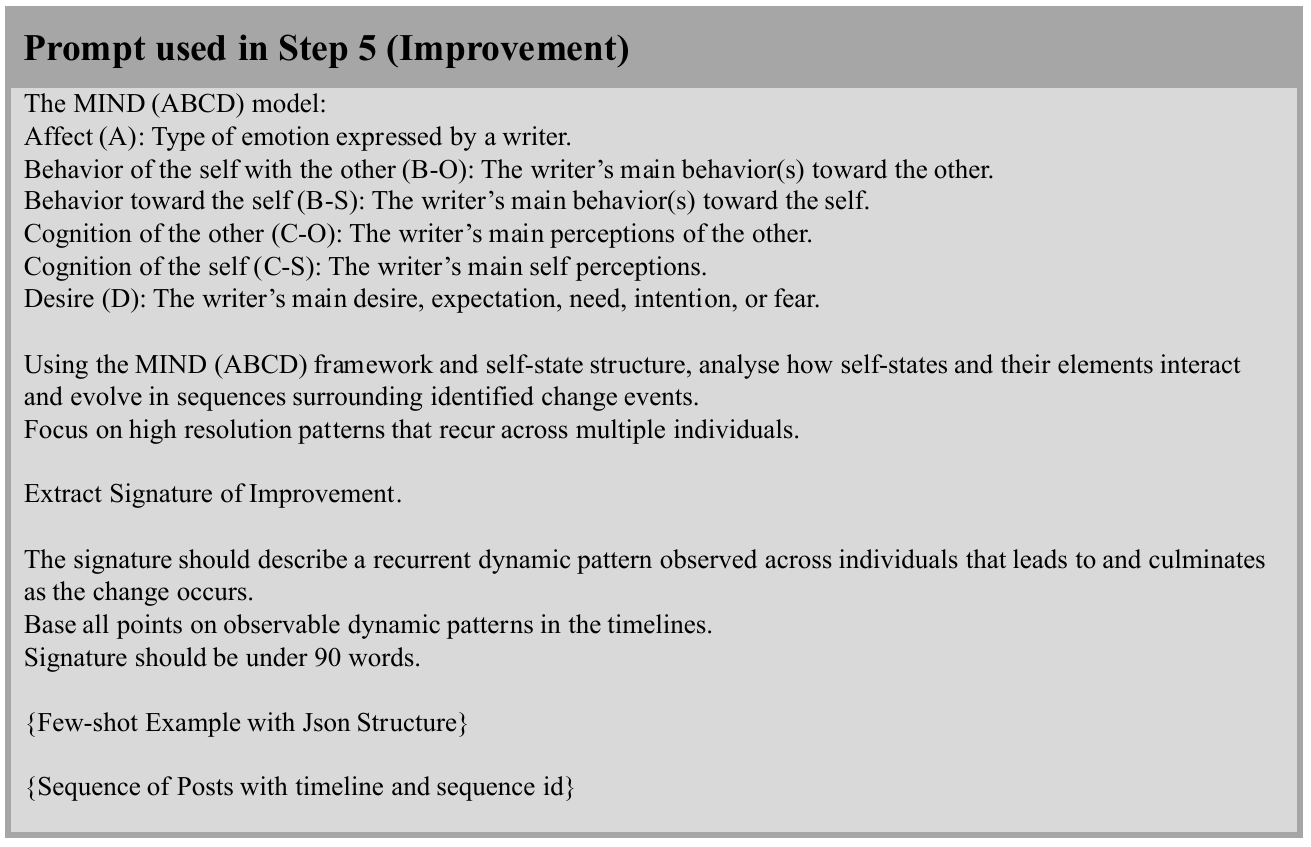}
    \caption{Prompt used for Step 5 (Improvement)}
    \label{fig:prompt_5_i}
\end{figure*}

\begin{figure*}[t]
    \centering
    \includegraphics[width=1.0\linewidth]{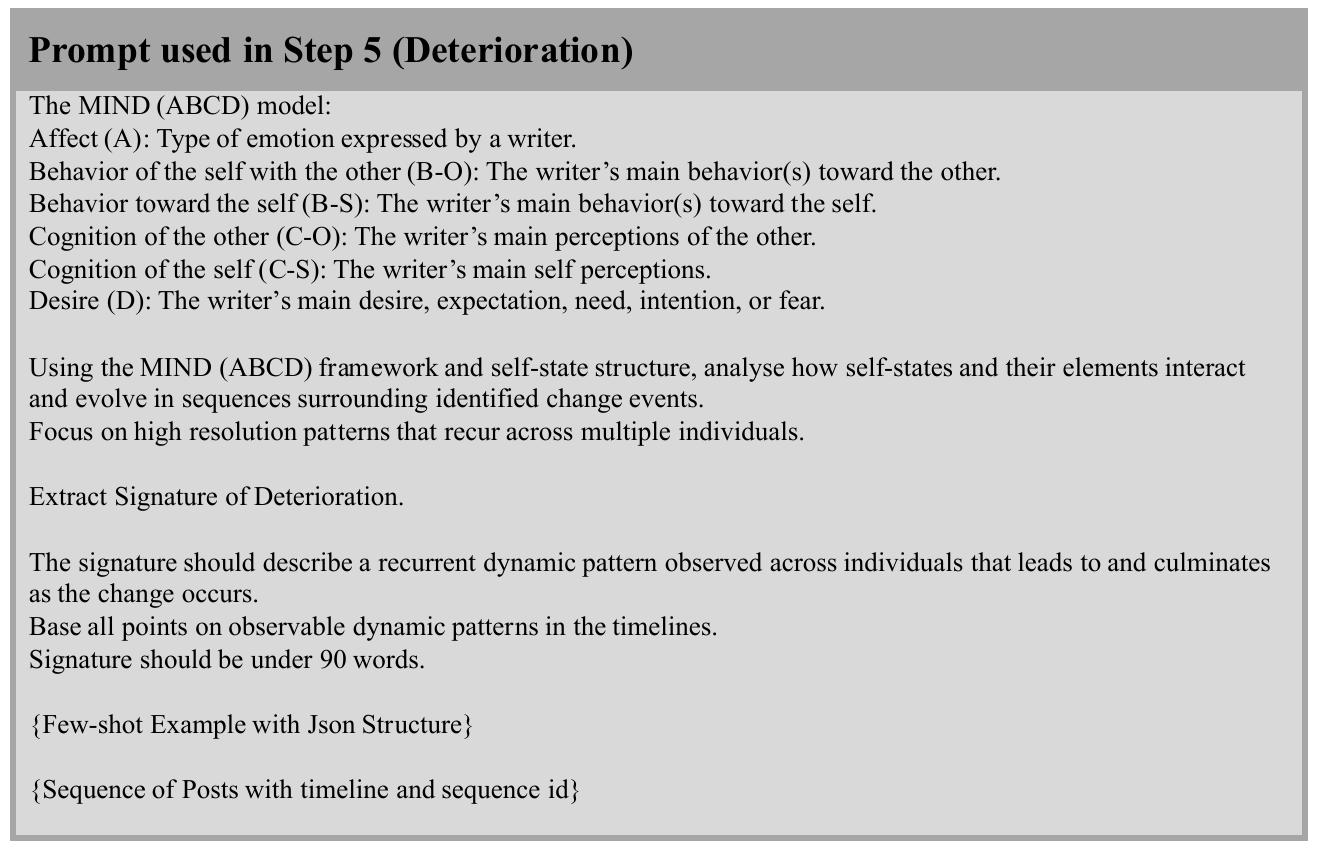}
    \caption{Prompt used for Step 5 (Deterioration)}
    \label{fig:prompt_5_d}
\end{figure*}

\km{As the final step, building on the MIND (ABCD) framework and the self-state structure, we analyze how self-state components interact and evolve across sequences surrounding change events, with the aim of identifying and summarizing dynamic patterns of psychological deterioration and improvement that recur across individuals. To this end, we perform this analysis in a zero-shot manner by prompting Qwen3.5-\hj{9}B~\cite{qwen3_5}. The prompt is provided in Figure~\ref{fig:prompt_5_i} and Figure~\ref{fig:prompt_5_d}.}
\hj{Concatenating every given summary at once as the model input resulted in Out-of-Memory error. We detoured this error by first splitting the given sequences into groups, extracting improvement and deterioration signature from each, and finally summarizing the signatures.}
\section{Experiment}

\subsection{Experimental Setting}

\km{In this section, we describe the experimental settings used across each step. For Steps 1, 2, and 3, we employed Qwen Embed as the backbone model and fine-tuned it using LoRA, with the rank $r = 16$ and scaling factor $\alpha = 32$. The learning rate was set to $1 \times 10^{-5}$, and a K-fold cross-validation strategy ($\text{K} = 5$) was applied consistently across all three steps. The maximum input token length was set to 786, the batch size to 16, and all models were trained for 30 epochs. \hh{ All experiments were conducted on a single NVIDIA RTX A6000 GPU, and the random seed was fixed to 42 for all experiments. Training a single fold took approximately 42 min for Step 1 and 15 min for Step 2, with the K-fold ensemble scaling roughly linearly.} For the SFT in Step 4, we employed LoRA with a rank of $r=8$ and scaling factor $\alpha=16$, attaching adapters to all linear layers. Training was performed with a learning rate of 2e-4, a batch size of 8, for 10 epochs.}

\begin{table}[t]
\centering
\small
\setlength{\tabcolsep}{5pt}
\begin{tabular}{cccc}
\toprule
 & & Step 1 & Step 2 \\
\cmidrule(lr){3-3} \cmidrule(lr){4-4}
Adaptive & K-fold & Macro F1 ($\uparrow$) & RMSE ($\downarrow$) \\
\midrule
\multirow{3}{*}{$\times$}
 & None   & 0.232 & 0.784 \\
 & Mean   & 0.210 & 0.677 \\
 & Voting & 0.210 & \textbf{0.612} \\
\midrule
\multirow{3}{*}{$\checkmark$}
 & None   & \textbf{0.333} & 0.777 \\
 & Mean   & 0.312 & 0.685 \\
 & Voting & 0.332 & 0.700 \\
\bottomrule
\end{tabular}
\caption{Results on the validation set for Step~1 and Step~2. \textit{Adaptive} indicates whether class-wise adaptive weighting is applied to the classification loss. For K-fold, ``None'' denotes a single model trained without K-fold, ``Mean'' aggregates the logits of the ensembled models before selecting the final prediction, and ``Voting'' determines the final prediction by majority voting.}
\label{tab:task1}
\end{table}

\begin{table}[t]
\centering
\small
\setlength{\tabcolsep}{5pt}
\begin{tabular}{lccc}
\toprule
Task & Adaptive & K-fold & Score \\
\midrule
\multirow{2}{*}{Step~1 (F1 $\uparrow$)}
 & $\checkmark$ & None & 0.320          \\
 & $\checkmark$ & Mean & \textbf{0.361} \\
\midrule
\multirow{3}{*}{Step~2 (RMSE $\downarrow$)}
 & $\checkmark$ & None   & 1.044          \\
 & $\times$     & Mean   & 1.010          \\
 & $\times$     & Voting & \textbf{1.003} \\
\bottomrule
\end{tabular}
\caption{Test-set results for Step~1 and Step~2.
Other conventions follow Table~\ref{tab:task1}.}
\label{tab:task1_test}
\end{table}

\subsection{Results}

\paragraph{Performance on Step 1 and Step 2:} 

\km{Prior to evaluation on the test set, we randomly partition the full training data into training and validation splits at an 80/20 ratio. Table~\ref{tab:task1} reports the subelement-average macro F1 (Step 1) and RMSE (Step 2) on the held-out validation set. For both Step~1 and Step~2, we examine two design factors: i) whether adaptive weighting is applied to the classification loss (denoted by the \textit{Adaptive} column), and ii) the K-fold aggregation strategy.}

\km{For \textbf{Step~1}, adaptive weighting consistently outperforms its non-weighted counterpart across all K-fold configurations, suggesting that adaptive weighting effectively mitigates the class imbalance inherent in the label distribution (Table~\ref{tab:task1}). In contrast, once adaptive weighting is applied, the choice of K-fold aggregation strategy (mean or majority voting) yields only marginal differences relative to the gain obtained from adaptive weighting itself (Table~\ref{tab:task1}). However, when these K-fold strategies are applied to the test set (Table~\ref{tab:task1_test}), ensembling provides a clear improvement over the single-model baseline. We conjecture that this discrepancy stems from the difference in data utilization: the single-model setting (K-fold = None) is trained on only 80\% of the training data, as the remaining 20\% is held out for model selection, whereas the K-fold ensemble leverages the entire training set across its folds, \hj{generalizing better} on the test set.}

\km{For \textbf{Step~2}, we observe that disabling adaptive weighting yields stronger validation performance (Table~\ref{tab:task1}). This trend is preserved on the test set (Table~\ref{tab:task1_test}), where the unweighted configuration again outperforms its weighted counterpart. A notable divergence from the validation results, however, is that K-fold aggregation produces consistent gains at test time across both aggregation strategies, where the \emph{mean} strategy averages the continuous outputs of the per-fold models and the \emph{voting} strategy rounds each fold's prediction to the nearest integer score and selects the mode as the final prediction. We attribute this improvement to the more complete utilization of the training data afforded by K-fold ensembling, which in turn translates into stronger generalization at test time.}

\paragraph{Performance on Step 3:} \km{Table~\ref{tab:task2} presents the prediction performance on the Detection of Moments of Change (MOC). As shown in the table, the 4B model outperforms the 8B model, which is likely attributable to the limited size of the training data (approximately 500 samples). With such a small dataset, the larger model is more prone to overfitting and fails to fully leverage its capacity, resulting in degraded performance. Furthermore, ensembling the predictions of five models trained via K-fold cross-validation on top of Qwen3-4B yields additional performance gains. This suggests that maximally utilizing all available training data through K-fold ensemble is beneficial, particularly in few-shot learning settings.} \hh{For the same reason, we kept the sliding window at size 2. Although a longer context could in principle capture Escalation more faithfully, models consuming wider temporal spans tend to overfit under our limited training regime, mirroring the overfitting pattern observed at the model-capacity level.}

\begin{table}[t]
\centering
\small
\renewcommand{\arraystretch}{1.15}
\setlength{\tabcolsep}{12pt}
\begin{tabular}{lc}
\toprule
\textbf{Model} & \textbf{F1 Score} ($\uparrow$) \\
\midrule
Qwen3-4B & 0.53 \\
Qwen3-8B & 0.49 \\
\midrule
Qwen3-4B w/ K-fold & \textbf{0.55} \\
\bottomrule
\end{tabular}
\caption{Performance on the test set of Step 3.}
\label{tab:task2}
\end{table}

\paragraph{Performance on Step 4 and Step 5:} \km{Table~\ref{tab:task3_1} presents the performance on Step 4. As shown in the table, Qwen3.5 generally achieves higher performance than Qwen3 across most metrics. However, we observe that the Contradiction score of Qwen3.5 is worse than that of Qwen3. Nevertheless, both models underperform the Baseline, which relies on zero-shot prompting. We attribute this outcome to two potential factors: 1) noise propagated from the results of the preceding Steps 1--3 may have degraded the final performance, and 2) the training dataset may have been too small relative to the model's capacity, suggesting that a more carefully designed prompting could have been necessary.} \hh{In such a low-resource regime, fine-tuning may also overwrite part of the LLM's pretrained knowledge, whereas the zero-shot baseline preserves it in full. This is consistent with our Step 3 finding that the smaller Qwen3-4B outperforms the larger Qwen3-8B (Table 3). }\km{For Step 5, we adopted a zero-shot approach, which resulted in performance scores of 0.7266 on Improvement and 0.2301 on Deterioration.}

\begin{table}[h]
\centering
\footnotesize
\renewcommand{\arraystretch}{1.15}
\setlength{\tabcolsep}{4pt}
\begin{tabular}{lccc}
\toprule
\textbf{Model} & \textbf{Consistency} & \textbf{Contradiction} & \textbf{Rouge-L} \\
               & ($\uparrow$)         & ($\downarrow$)           & \textbf{Recall} ($\uparrow$) \\
\midrule
Baseline   & 0.763          & 0.753          & 0.255 \\
\midrule
Qwen3.5-4B & \textbf{0.669} & 0.857          & \textbf{0.290} \\
Qwen3-4B   & 0.654          & \textbf{0.834} & 0.284 \\
\bottomrule
\end{tabular}
\caption{Performance on the test set of Step 4.}
\label{tab:task3_1}
\vspace{-3mm}
\end{table}


\section{Discussion} \label{sec:discuss}

\paragraph{Future Works}\km{Our analysis of the dataset provided for the shared task revealed a significant class imbalance across the sub-dimensions used in the Step 1 post-level classification. More importantly, several sub-dimensions contained only a single training instance, and some contained none at all. This severe sparsity likely limited the model's ability to learn reliable decision boundaries for these categories, and may partly explain the lower performance observed on low-resource labels. To mitigate this issue, we adopted a K-fold ensemble strategy in the present work. However a more direct solution, which we leave for future research, is to use LLMs to generate synthetic user posts for the underrepresented sub-dimensions. This approach could help reduce both the imbalance and the missing-label problems, leading to a more balanced training signal across all labels.}

\section{Conclusion}

\km{In this paper, we presented an LLM-based pipeline for the CLPsych 2026 shared task on capturing mental health changes through social media timelines. To address the class imbalance that commonly appears in mental health data, we designed a loss function that reflects the label distribution and combined it with a K-fold ensemble strategy to improve robustness. We also enabled the model to describe users' mental states in natural language, either through fine-tuning or few-shot examples, making the outputs easier to interpret. Building on these components, our pipeline brings post-level and user-level temporal modeling together into a single unified framework.}
\section*{Limitations}

\km{In this paper, we proposed a holistic pipeline for Capturing and Characterizing Mental Health Changes through Social Media Timeline Dynamics. Nevertheless, several limitations remain to be addressed in future work. First, the scarcity of training data poses a significant challenge. The dataset employed for training exhibits a substantial class imbalance, with only approximately 500 samples available in total. Consequently, models with larger capacities tend to suffer from severe overfitting. To mitigate this issue, one promising direction is to leverage LLM-based data augmentation to generate additional samples, thereby constructing a more balanced dataset across all classes. Second, the design of sophisticated prompts tailored to the mental health domain warrants further investigation. The performance of LLMs is known to be highly sensitive to subtle variations in prompt formulation. Although this aspect was beyond the scope of the present study, we believe that incorporating domain-specific knowledge related to mental health into prompt engineering could further enhance the effectiveness of the proposed pipeline.}

\section*{Ethics} The dataset provided for the CLPsych 2026 shared task contains sensitive content related to users' mental health. Throughout this work, we strictly followed the data handling rules announced by the CLPsych organizers and we avoided any external API service that would have required transmitting the data outside our local environment. Our five-step pipeline is intended as a research artifact for studying mental health dynamics, and any use of it for screening or diagnosis would require oversight by trained mental health professionals. Automated inference of self-states and moments of change carries non-trivial clinical implications when errors occur: false positives may lead to unnecessary intervention, while false negatives could mean a missed opportunity for support.

Our proposed future direction of LLM-based synthetic data generation for underrepresented sub-dimensions (Section~\ref{sec:discuss}) also carries ethical risks that warrant careful consideration. LLMs trained on broad web data may misrepresent the lived experience of clinical populations and produce stereotypical or inaccurate depictions of distress, and any such distortions may be silently propagated once the generated samples are incorporated into training. Generating content in sensitive categories such as self-harm further raises content-safety concerns. Before using such generated data for training, future work should have clinical experts review the samples and check for fairness across different demographic and language groups.


\section*{Acknowledgments}

This work was supported by the Korean Government through the grants from IITP (RS-2021-II211343, RS-2025-25442338, 26-InnoCORE-01)

\bibliography{custom}

\appendix

\section{Appendix}
\label{sec:appendix}

\subsection{Related Works}

\subsubsection{NLP Approaches in Psychological Assessment}

\km{The proliferation of large-scale datasets and the rapid advancement of computational resources have driven the emergence of foundation models capable of performing a wide range of tasks in a generalized manner, spanning both computer vision and natural language processing. These breakthroughs have subsequently inspired researchers to explore the applicability of such models in specialized domains, including psychology and mental healthcare~\cite{tsakalidis-etal-2022-overview, atzil_slonim_2026}. MentalBERT~\cite{ji-etal-2022-mentalbert} represents an early effort in this direction, extending the general-purpose BERT architecture~\cite{devlin2019bert} through continued domain-adaptive pretraining on mental health-related corpora, enabling the extraction of psychologically grounded contextual embeddings. Building upon this line of work, MentalLLaMA~\cite{yang2024mentallama} shifts from a purely representational paradigm to a generative one by leveraging LLaMA~\cite{touvron2023llama}, an instruction-tuned large language model, and further applying supervised fine-tuning on a curated collection of mental health-specific instruction datasets. This allows MentalLLaMA to perform a diverse array of mental health tasks in a conversational and interpretable manner, going beyond the embedding-focused capabilities of its predecessor. Motivated by these advances in NLP for mental health, this paper aims to leverage large language models (LLM) trained on large-scale data to capture and characterize longitudinal changes in mental health states within social media timelines.}

\subsubsection{Collecting Psychological Data for NLP}

\km{Unlike general-purpose tasks, domains such as psychology present unique challenges in data collection, including privacy concerns and ethical constraints, which inevitably limit the scale and accessibility of psychological data. To address these limitations and advance NLP research in the psychological domain, \citeauthor{tsakalidis-etal-2022-identifying} curated a large-scale mental health counseling dataset sourced from social media. Building upon this dataset, subsequent works have explored a diverse range of mental health-related tasks, including mood change detection, suicide risk classification, and the identification and summarization of suicide risk evidence, thereby contributing to the broader development of AI for psychology. Motivated by these advances, this paper presents a framework proposed as part of the CLPsych shared task, aimed at capturing and characterizing mental health changes through social media timeline dynamics.}

\subsection{Prompt Specifications}

\begin{figure*}[t]
    \centering
    \includegraphics[width=1.0\linewidth]{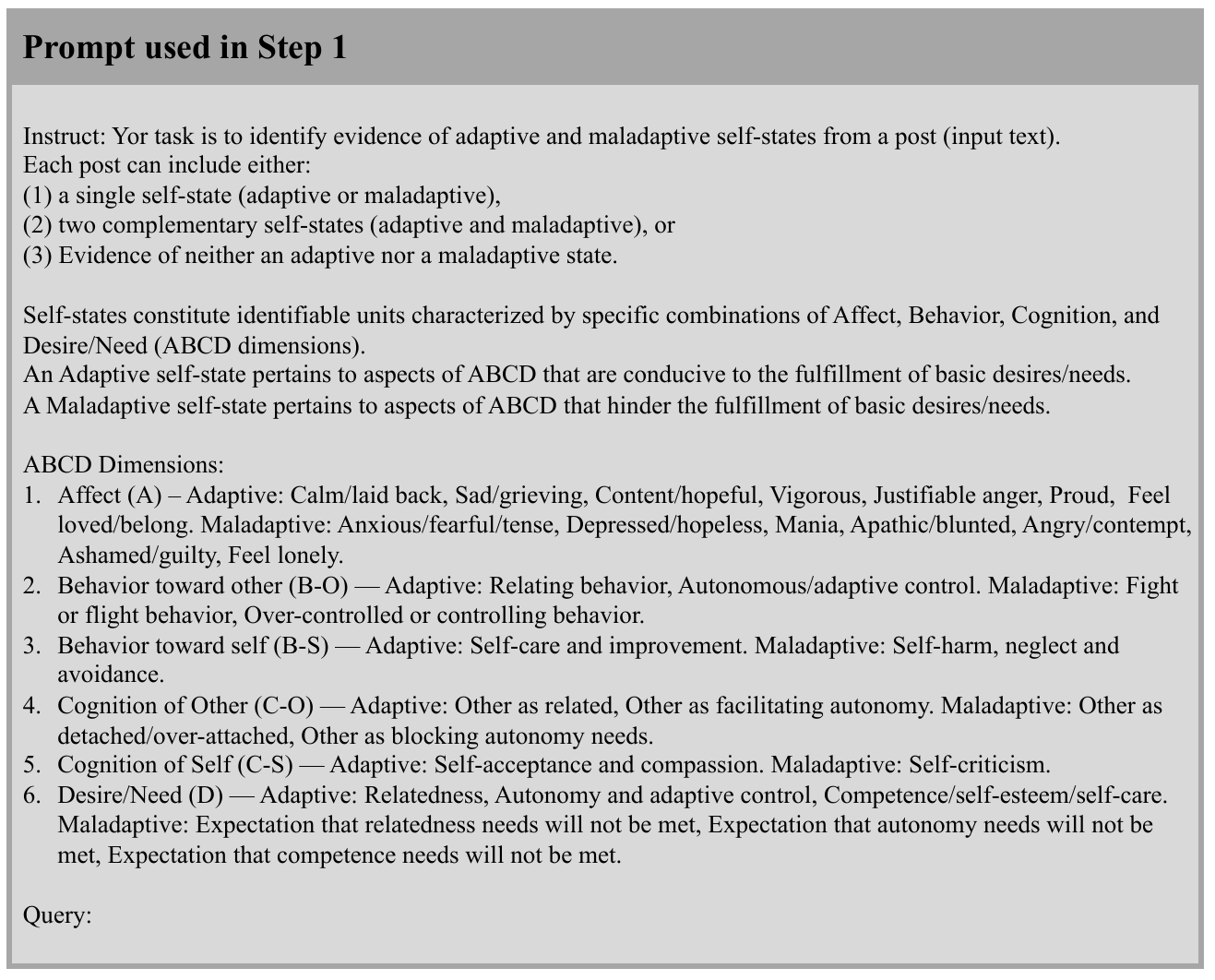}
    \caption{Prompt used for Post-Level Identification of Dominant ABCD Sub-elements and Self-State Composition}
    \label{fig:prompt_1}
\end{figure*}

\begin{figure*}[t]
    \centering
    \includegraphics[width=1.0\linewidth]{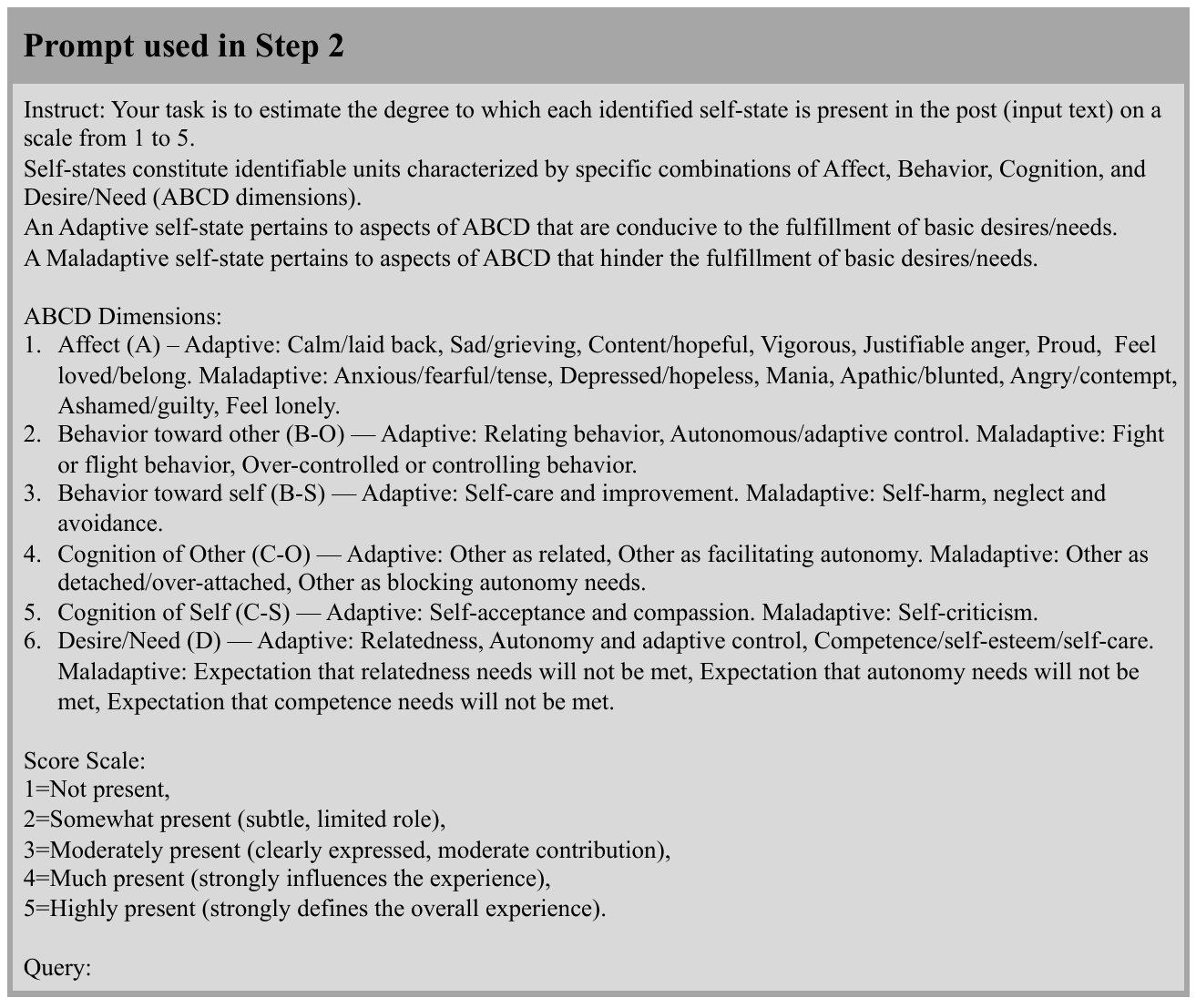}
    \caption{Prompt used for Self State Presence Rating}
    \label{fig:prompt_2}
\end{figure*}

\begin{figure*}[t]
    \centering
    \includegraphics[width=1.0\linewidth]{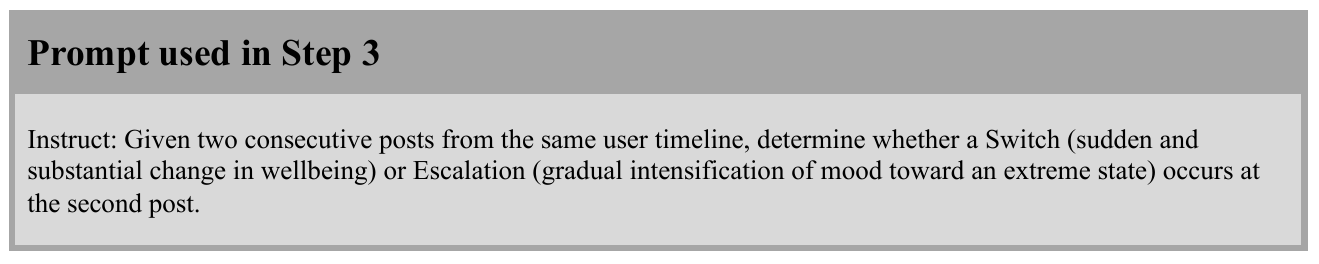}
    \caption{Prompt used for Identify Moments of Change (MOC)}
    \label{fig:prompt_3}
\end{figure*}

\begin{figure*}[t]
    \centering
    \includegraphics[width=1.0\linewidth]{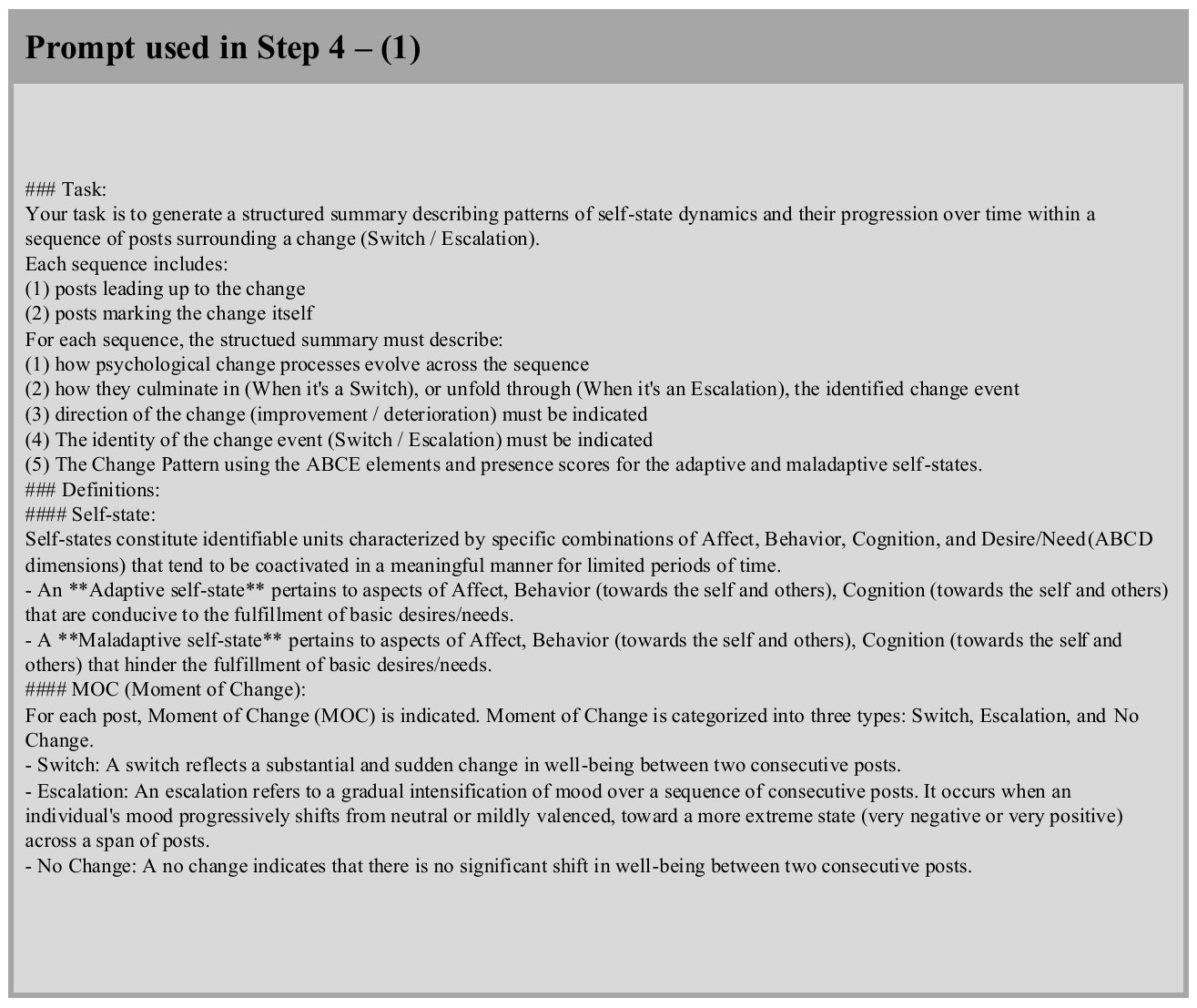}
    \caption{Prompt used for Summarization}
    \label{fig:prompt_4_1}
\end{figure*}

\begin{figure*}[t]
    \centering
    \includegraphics[width=1.0\linewidth]{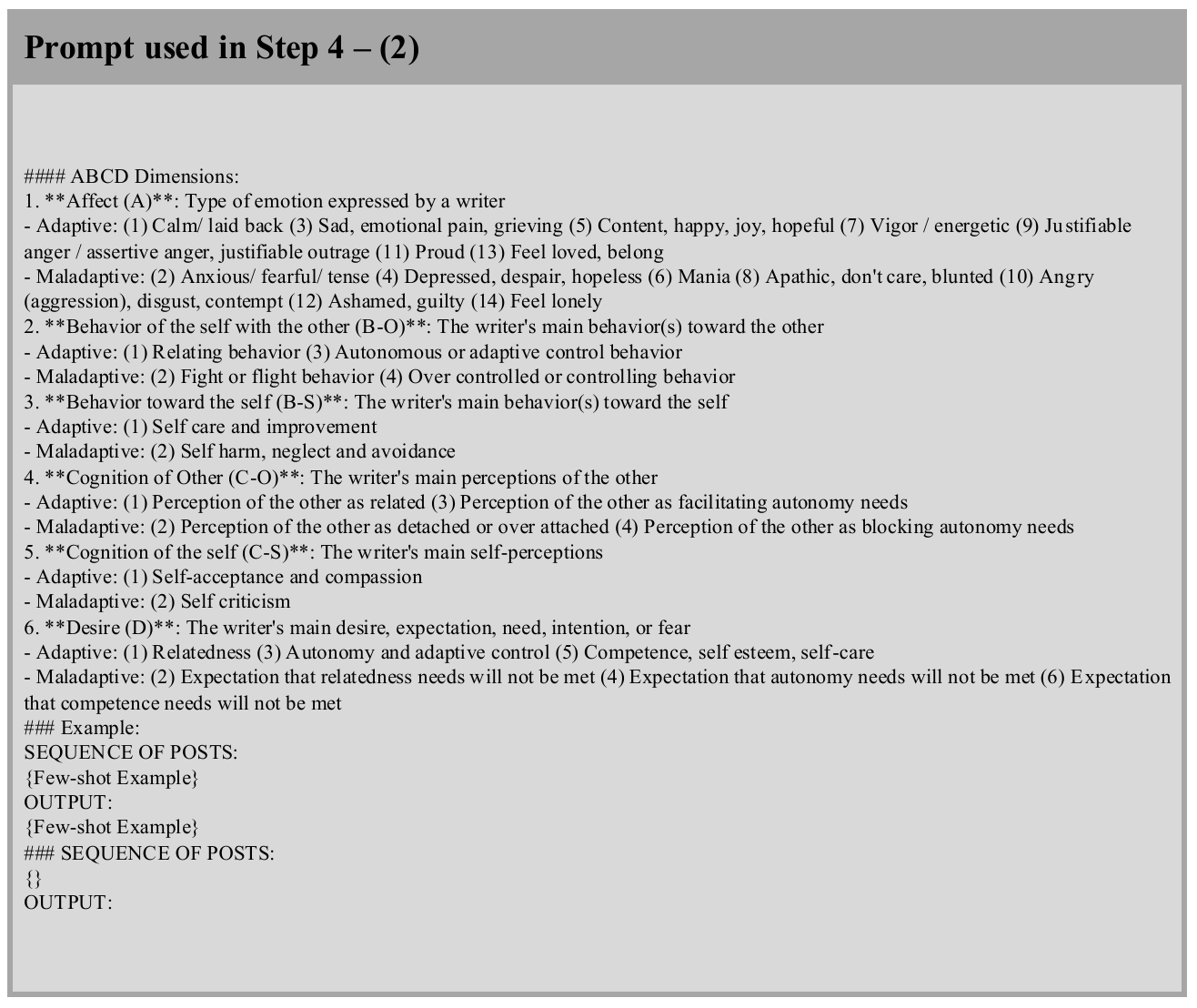}
    \caption{Prompt used for Summarization}
    \label{fig:prompt_4_2}
\end{figure*}

\km{Figures~\ref{fig:prompt_1},~\ref{fig:prompt_2},~\ref{fig:prompt_3},~\ref{fig:prompt_4_1}, and~\ref{fig:prompt_4_2} illustrate the prompts employed in each respective task. For the few-shot examples in the Step~4 prompt, instances drawn from the provided dataset were utilized. Specifically, each example was constructed by incorporating the MoC label (\textit{i.e.,} Switch and Escalation indicators), the Maladaptive presence score, the Adaptive presence score, and the composition of each corresponding state as input.}

\end{document}